\documentclass[letterpaper]{article} 
\usepackage{aaai2026}  
\usepackage{times}  
\usepackage{helvet}  
\usepackage{courier}  
\usepackage[hyphens]{url}  
\usepackage{graphicx} 
\usepackage{natbib}  
\usepackage{caption} 
\usepackage{algorithm}
\usepackage{algorithmic}
\usepackage[table]{xcolor}
\usepackage{pifont}

\usepackage{amsmath}

\usepackage{newfloat}
\usepackage{listings}
\DeclareCaptionStyle{ruled}{labelfont=normalfont,labelsep=colon,strut=off} 
\floatstyle{ruled}
\newfloat{listing}{tb}{lst}{}
\floatname{listing}{Listing}
\nocopyright 
\usepackage[bottom,flushmargin]{footmisc}   

\title{MapKD: Unlocking Prior Knowledge with Cross - Modal Distillation\\for Efficient Online HD Map Construction}
\author{
    Ziyang Yan$^{1}$\thanks{Equal contribution.},
    Ruikai Li$^{1}$\footnotemark[1]
    Zhiyong Cui$^{1}$\thanks{Corresponding author.},
    Bohan Li$^{3,4}$,
    Han Jiang$^{1}$,\\
    Yilong Ren$^{1}$,
    Aoyong Li$^{1}$,
    Zhenning Li$^{5}$,
    Sijia Wen$^{2}$,
    Haiyang Yu$^{1}$
}
\affiliations{
    \textsuperscript{\rm 1}State Key Laboratory of Intelligent Transportation System, Beihang University \\
    
    \textsuperscript{\rm 2}School of Aritificial Intelligence, Beihang University \\
    \textsuperscript{\rm 3}Shanghai Jiao Tong University 
    \textsuperscript{\rm 4}Ningbo Institute of Digital Twin, Eastern Institute of Technology \\
    \textsuperscript{\rm 5}State Key Laboratory of Internet of Things for Smart City, University of Macau \\
    \{yanziyang, rickyli, zhiyongc\}@buaa.edu.cn \\

}

\usepackage{bibentry}

\begin{document}

\maketitle

\begin{abstract}

Online HD map construction is a fundamental task in autonomous driving systems, aiming to acquire semantic information of map elements around the ego vehicle based on real-time sensor inputs. Recently, several approaches have achieved promising results by incorporating offline priors such as SD maps and HD maps or by fusing multi-modal data. However, these methods depend on stale offline maps and multi-modal sensor suites, resulting in avoidable computational overhead at inference. To address these limitations, we employ a knowledge distillation strategy to transfer knowledge from multimodal models with prior knowledge to an efficient, low-cost, and vision-centric student model. Specifically, we propose \textbf{MapKD}, a novel multi-level cross-modal knowledge distillation framework with an innovative Teacher-Coach-Student (TCS) paradigm. This framework consists of: (1) a camera-LiDAR fusion model with SD/HD map priors serving as the teacher; (2) a vision-centric coach model with prior knowledge and simulated LiDAR to bridge the cross-modal knowledge transfer gap; and (3) a lightweight vision-based student model. Additionally, we introduce two targeted knowledge distillation strategies: Token-Guided 2D Patch Distillation (TGPD) for bird's eye view feature alignment and Masked Semantic Response Distillation (MSRD) for semantic learning guidance. Extensive experiments on the challenging nuScenes dataset demonstrate that MapKD improves the student model by +6.68 mIoU and +10.94 mAP while simultaneously accelerating inference speed. The code is available at: \url{https://github.com/2004yan/MapKD2026}.
\end{abstract}

   
\vspace{-2mm}
\section{Introduction}

Online HD map construction is a critical yet challenging task in autonomous driving systems, generating semantic maps from real-time sensor inputs to support vehicle planning and navigation. While vision-centric online HD map construction methods have gained considerable attention due to their cost-effectiveness and superior map freshness, these approaches~\cite{bao2022hdmap,yuan2023streammapnet,pan2020crossview} often suffer from limited 3D geometric awareness and occlusion problems, which compromise accuracy. In contrast, recent advances in multimodal fusion~\cite{li2022hdmapnet,liu2023vectormapnet,ding2023pivotnet,jiang2025toward,liao2022maptr,chen2022pqtransformer} and prior-injection~\cite{xiong2023neural, priorMap2023,gao2023satellite,yan2023superfusion,xia2023mvmap,kong2023mapvr} paradigms have demonstrated remarkable precision improvements. Notably, PMapNet~\cite{jiang2024pmapnet}—a representative framework—effectively integrates LiDAR and multi-view camera data in Bird's Eye View (BEV) space while incorporating prior knowledge from both Standard Definition (SD) maps and offline-collected HD maps, thereby enabling robust long-range mapping capabilities. However, these methods incur substantial computational overhead due to their reliance on complex models, additional sensor requirements, and offline map processing, ultimately limiting their scalability for deployment.
\begin{figure}[t]
    \centering
    \includegraphics[width=\linewidth]{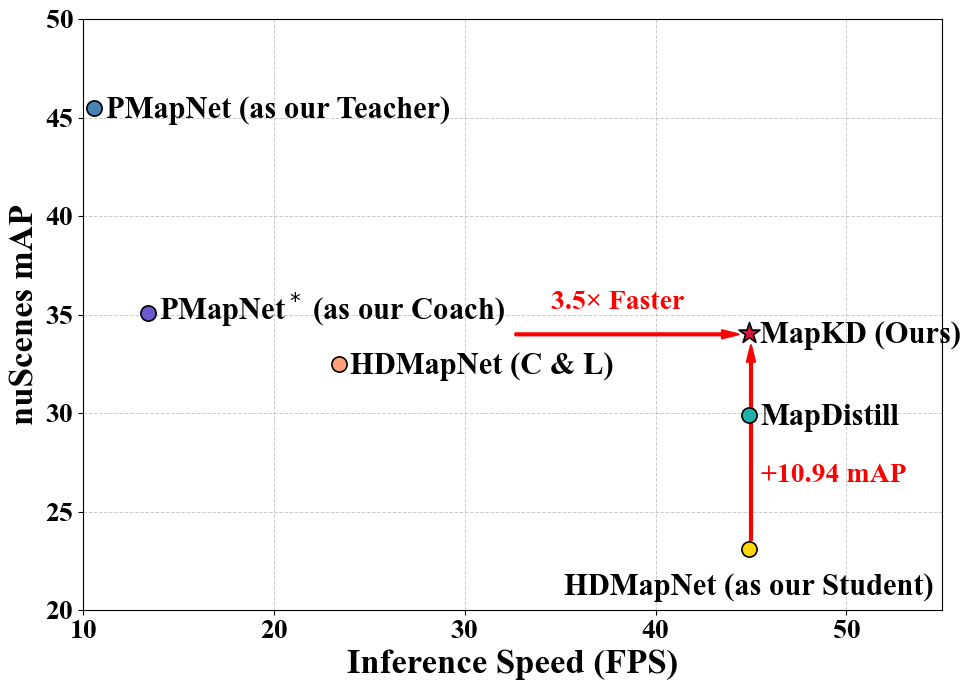}
    \caption{Comparison of our proposed MapKD with different methods on the nuScenes validation dataset.}
    \label{fig:comparison}
\end{figure}

\begin{figure*}[!t]
    \centering
    \includegraphics[width=\textwidth]{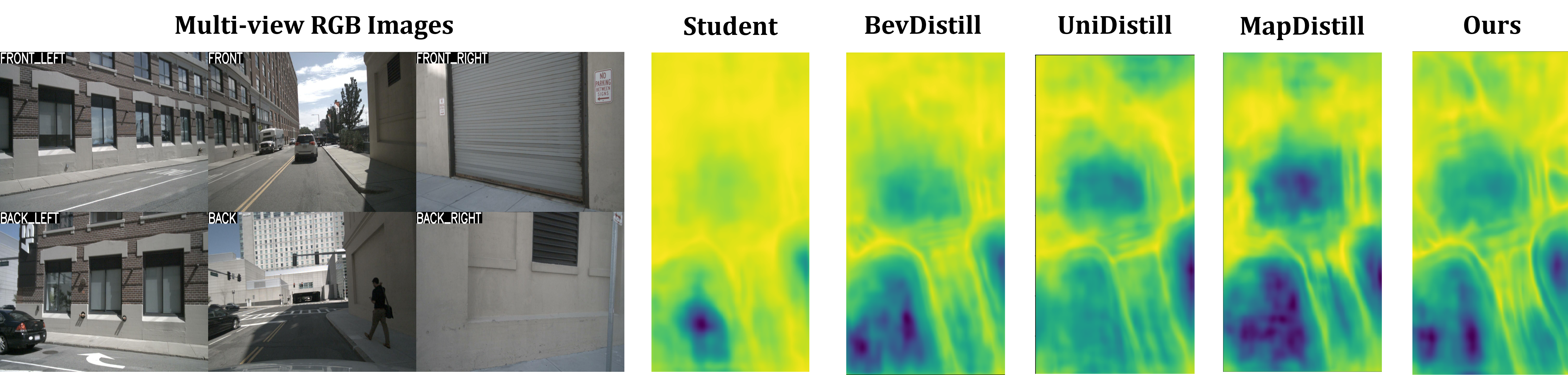}
    \caption{Qualitative comparison of BEV feature generation. Our MapKD produces more coherent and detailed BEV features than the student model without distillation (HDMapNet) and MapDistill in the representative scenario.}
    \label{fig:vis2}
\end{figure*}

To facilitate more efficient and scalable online HD map construction for autonomous driving systems, achieving an optimal balance between map performance and deployment cost remains critical. Knowledge distillation (KD) offers a well-established solution to achieve this goal by transferring knowledge from a larger, better-trained teacher model to a more compact student model. In the domain of autonomous driving, KD has been successfully applied to BEV-based perception frameworks. Specifically, some approaches~\cite{zhao2023simdistill,xu2024sckd,wang2024sparsekd,wang2023distillbev,lidar2map2023,yang2023lgkd,xu2024sckd}  have improved vision-centric compact models by distilling knowledge from stronger modalities — either camera-LiDAR fusion-based or LiDAR-only models — thereby achieving an effective balance between deployment cost and model accuracy.

Despite the promising results demonstrated by these cross-modal KD methods, several critical challenges remain unresolved: (1) Information loss across modalities: During knowledge transfer, the vision-centric student branch typically requires auxiliary pseudo-point cloud modules to compensate for the geometric information inherent in LiDAR data, which inevitably introduces additional computational overhead during inference~\cite{zhou2023unidistill,hao2024mapdistill}. (2) Limitations in prior knowledge transfer: Most existing KD frameworks employ nearly identical BEV post-processing architectures, which not only hinder the effective transfer of prior knowledge but also constrain the broader application of KD technology in autonomous driving systems.

To address these challenges, we propose \textbf{MapKD}, a novel teacher-coach-student knowledge distillation framework. This approach enables vision-centric lightweight student models to acquire cross-modal knowledge and prior information from both SD and HD maps through multi-level cross-modal alignment and distillation, facilitated by the coordinated integration of teacher and coach models.
Different from prior KD methods, MapKD introduces a modality-aligned coach network that bridges the gap between a full-modality teacher and a lightweight student. The coach is trained using extra simulated LiDAR features and structured map priors, enabling the student to learn spatial semantics with improved stability and generalization. We further propose two innovative distillation strategies: (1) Token-Guided Patch Distillation (TGPD) to align BEV representations via patch-token attention, and (2) Masked Semantic Response Distillation (MSRD) to refine semantic predictions on foreground regions using soft supervision. 

Extensive experiments on the nuScenes dataset demonstrate the effectiveness and efficiency of our approach. Not only does our MapKD achieve significant improvements over the camera-only student baseline, it furthermore matches the performance of the coach model with map priors, while running inference 3.5 times faster than the coach as shown in Figure~\ref{fig:comparison}. These results highlight the advantage of our three-stage design and cross-modal knowledge transfer.

To summarize, our main contributions are as follows:
\begin{itemize}
    \item This study proposes \textbf{MapKD}, which leverages multi-stage knowledge distillation to transfer knowledge from  multimodal models with priors to a lightweight student that does not depend on such priors, thus reducing the need for expensive HD map pre-training.
    \item This study proposes a novel Teacher-Coach-Student (TCS) distillation framework that introduces a simulated LiDAR coach network by our innovatively proposing. The coach-involved design bridges the modality and capacity gap between the full-modality teacher and the lightweight camera-only student, enabling more effective knowledge transfer.
    \item This study designs two targeted and complementary distillation strategies: Token-Guided 2D Patch Distillation (TGPD) for BEV feature alignment via patch-level token interactions, and Masked Semantic Response Distillation (MSRD) for foreground-aware semantic supervision. These strategies jointly enhance both spatial and semantic learning. Our method demonstrates enhanced BEV feature quality compared to existing approaches, as shown in Figure~\ref{fig:vis2}, and achieves a performance gain of 
    \textbf{+6.68 mIoU} and \textbf{+10.94 mAP} over the baseline camera-only student.
\end{itemize}

\begin{figure*}[!t]
    \centering
    \includegraphics[width=\linewidth]{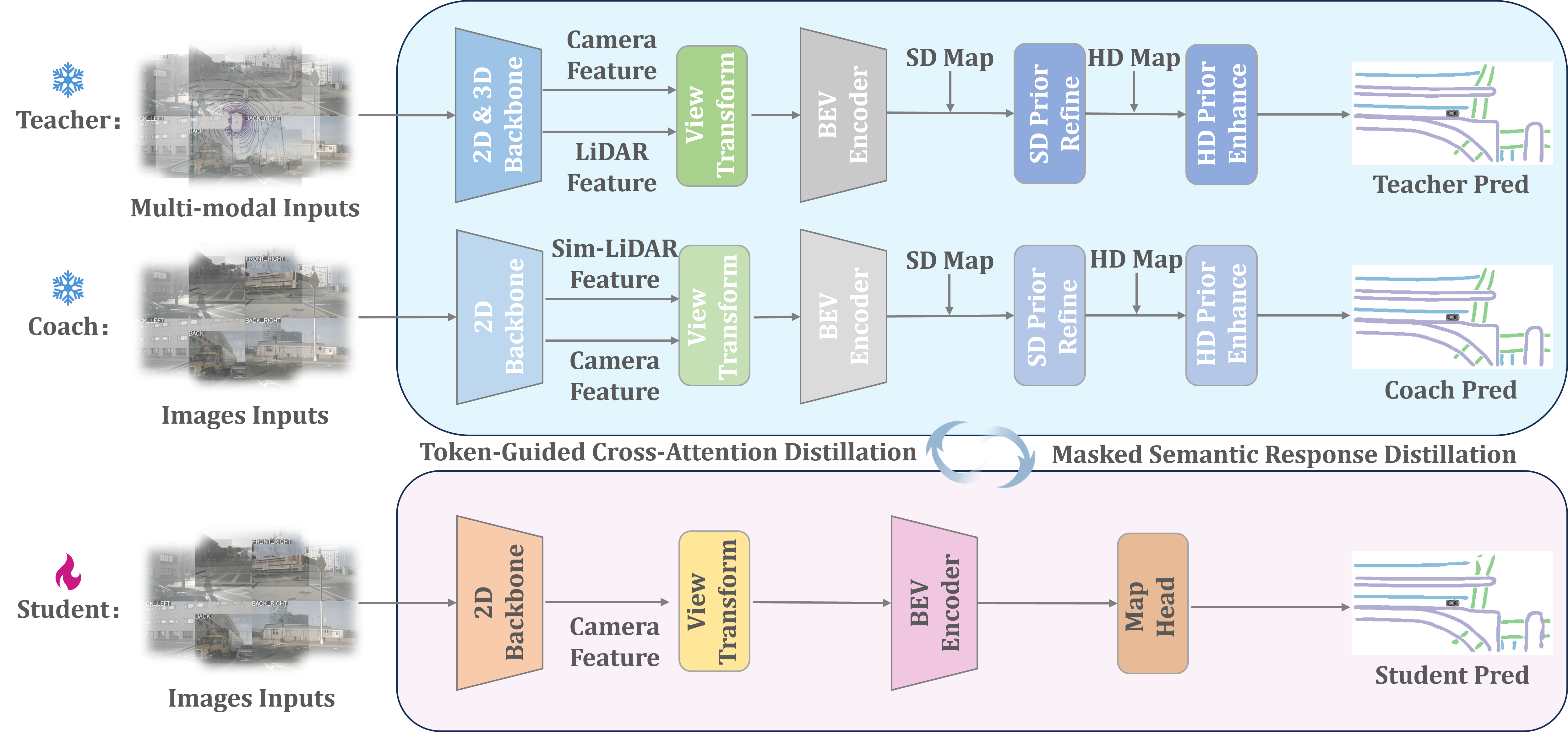}
    \caption{The Overview of our MapKD Framework. The teacher provides full-modality supervision; the coach bridges the modality gap; and the student learns to generate HD maps using camera input only. During inference, only the student model is used.}
    \label{fig:tcs_framework}
\end{figure*}

\vspace{-2mm}
\section{Methodology}
\subsection{Teacher-Coach-Student Distillation Framework}

An overview of our proposed Teacher-Coach-Student Distillation Framework is illustrated in Figure~\ref{fig:tcs_framework}. Our distillation framework consists of three models: a full-modality teacher $\mathcal{T}$, a modality-aligned coach $\mathcal{C}$ through our unique proposal, and a camera-only student $\mathcal{S}$. Each plays a distinct role in the training pipeline, enabling multi-level supervision and progressive modality alignment.

\paragraph{Teacher Model $\mathcal{T}$.}
The teacher model is based on PMapNet ~\cite{jiang2024pmapnet}. It effectively integrates LiDAR and map priors,    providing strong multimodal information. It receives multi-view images $I$, LiDAR point cloud $L$, SD map $M_{\mathrm{SD}}$, and HD map $M_{\mathrm{HD}}$. The processing pipeline is shown in the first line of Figure~\ref{fig:tcs_framework} and described as follows.

First, image features and LiDAR features are extracted through modality-specific encoders:
\begin{equation}
F_I = \text{Enc}_{\mathrm{img}}(I), \quad F_L = \text{Enc}_{\mathrm{lidar}}(L) 
\end{equation}

These two features are concatenated and transformed into BEV space:
\begin{equation}
F_{\mathrm{T}}^{\mathrm{BEV}} = \text{BEVProj}(\text{Concat}(F_I, F_L)) 
\end{equation}

Subsequently, BEV features are fused with both SD and HD map priors in sequence, yielding a unified representation:
\begin{equation}
F_{\mathrm{BEV}}^{\mathrm{HD}} = \text{HDFusion}(\text{SDFusion}(F_{\mathrm{T}}^{\mathrm{BEV}}, M_{\mathrm{SD}}), M_{\mathrm{HD}})
\end{equation}

Finally, the semantic decoder outputs the predicted HD map logits:
\begin{equation}
S_T = \text{Dec}_{\mathrm{sem}}(F_{\mathrm{BEV}}^{\mathrm{HD}}) 
\end{equation}

The resulting teacher prediction $S_T$ and intermediate BEV features $F_{\mathrm{BEV}}^{\mathrm{HD}}$ are used as distillation targets for downstream modules.

\paragraph{Coach Model $\mathcal{C}$.}
Coach employs the same overall architecture as the teacher but replaces real LiDAR with simulated LiDAR features generated from camera input. Specifically, the input consists of multi-view images $I$ and map priors $M_{\mathrm{SD}}$ and $M_{\mathrm{HD}}$. The processing pipeline is shown in the second line of Figure~\ref{fig:tcs_framework} and described as follows.

First, the image $I$ is processed through two separate encoders:
\begin{equation}
\tilde{F}_L = \text{Enc}_{\mathrm{3D}}(I), \quad F_I = \text{Enc}_{\mathrm{2D}}(I) 
\end{equation}
where $\tilde{F}_L$ is the simulated LiDAR features produced by a 3D encoder \cite{philion2020lift}, and $F_I$ is the 2D image feature.

The two features are concatenated and projected into BEV space:
\begin{equation}
F_{\mathrm{C}}^{\mathrm{BEV}} = \text{BEVProj}(\text{Concat}(F_I, \tilde{F}_L)) 
\end{equation}

Then, BEV features are fused with both SD and HD map priors in sequence:
\begin{equation}
F_{\mathrm{BEV}}^{\mathrm{HD}} = \text{HDFusion}(\text{SDFusion}(F_{\mathrm{C}}^{\mathrm{BEV}}, M_{\mathrm{SD}}), M_{\mathrm{HD}})
\end{equation}

The semantic decoder ultimately generates the logit output:
\begin{equation}
S_C =  \text{Dec}_{\mathrm{sem}}(F_{\mathrm{BEV}}^{\mathrm{HD}})
\end{equation}

The coach model serves to provide modality-consistent intermediate guidance to the student.

\paragraph{Student Model $\mathcal{S}$.}
The student model is based on HDMapNet ~\cite{li2022hdmapnet}. It is lightweight and camera-only, making it suitable for real-time deployment.
The processing pipeline is shown in the third line of Figure~\ref{fig:tcs_framework} and described as follows.

First, the input image is encoded to extract 2D features:

\begin{equation}
F_I = \text{Enc}_{\mathrm{2D}}(I) 
\end{equation}

These features are then projected into BEV space using a learned inverse perspective mapping (IPM) module:
\begin{equation}
F_{\mathrm{S}}^{\mathrm{BEV}} = \text{BEVProj}(F_I) 
\end{equation}

Lastly, the BEV features are decoded to produce the predicted HD map logits:
\begin{equation}
S_S = \text{Dec}_{\mathrm{sem}}(F_{\mathrm{S}}^{\mathrm{BEV}}) 
\end{equation}
\begin{figure}[H] 
    \centering
    \includegraphics[width=0.5\textwidth]{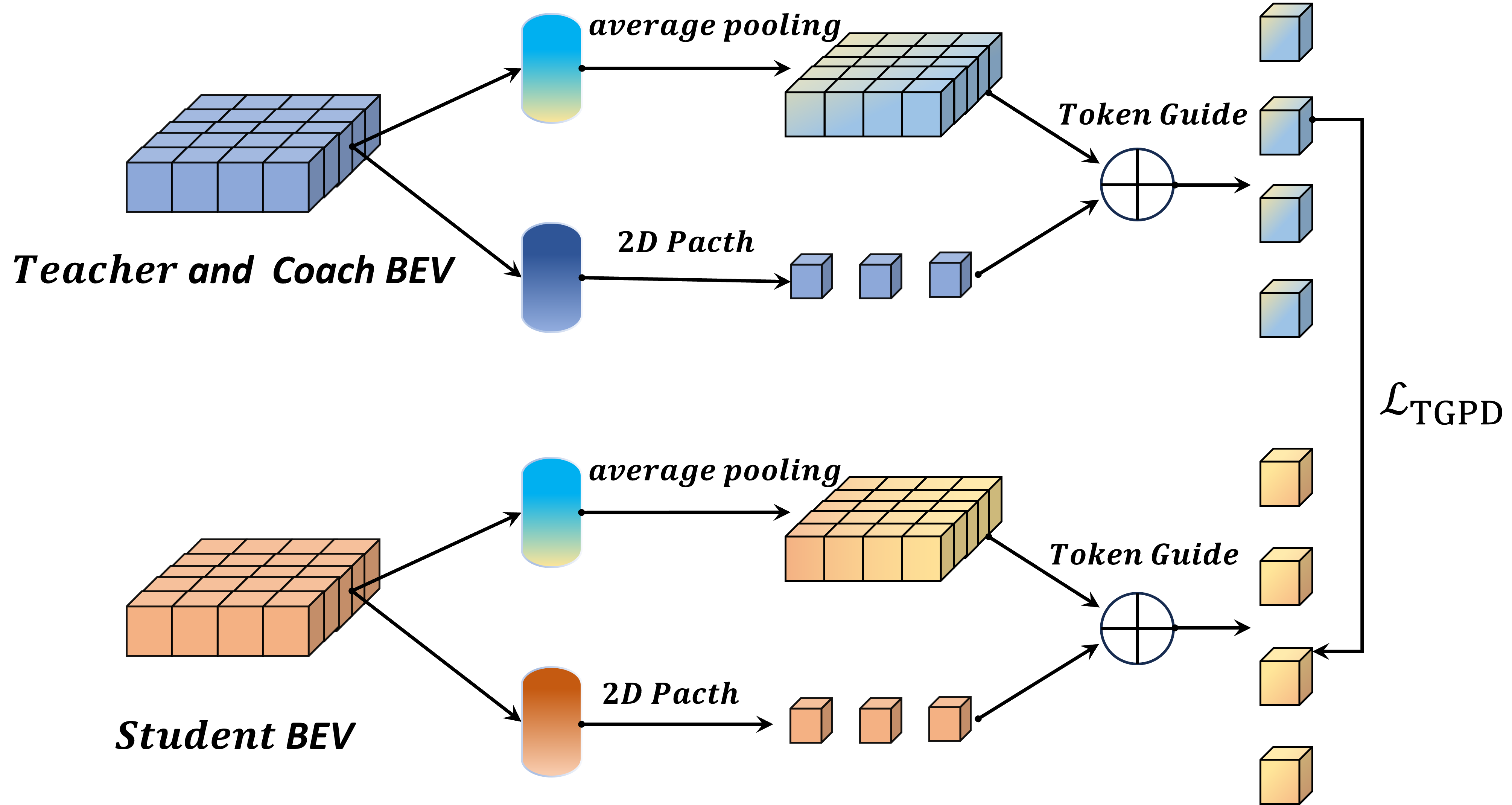} 
    \caption{Illustration of the proposed Token-Guided 2D Patch Distillation (TGPD) process.} 
    \label{fig:TGPD}
\end{figure}

The student is trained under both supervised losses and distillation signals from the teacher $\mathcal{T}$ and coach $\mathcal{C}$.
\subsection{Distillation Strategy Design}
To enhance the student model’s learning capacity at both the feature and output levels, we introduces the  two specific distillation methods: TGPD and MSRD.

\subsubsection{Token-Guided 2D Patch Distillation (TGPD).}

As shown in Figure~\ref{fig:TGPD}, Token-Guided 2D Patch Distillation (TGPD) is proposed to enhance the student model’s BEV feature learning by incorporating both local details and global context.

First, the BEV feature maps from the student, teacher and coach are divided into small 2D patches, facilitating local feature comparison. Each patch is then converted into a vector through a linear projection.
Simultaneously, the average of the entire BEV feature map is computed to obtain a global token, which summarizes the overall scene. This token is projected into the same embedding space as the patch vectors.

The global token and all patch vectors are then concatenated into a single sequence, over which attention is computed to capture the interactions between global and local information.

Finally, the student’s attention results are aligned with those of the teacher and coach. KL divergence is employed to match the attention distributions, while mean squared error (MSE) is used to align the corresponding feature values.
The distillation loss from the teacher to the student is defined as:
\begin{equation}
\mathcal{L}^{\text{TG}}_{\text{T2S}} = \text{KL}\left( \frac{A^{\text{stu}}}{\tau} \bigg\| \frac{A^{\text{tea}}}{\tau} \right) + \lambda \cdot \left\|F_{\mathrm{BEV}}^{\mathrm{S}} - F_{\mathrm{BEV}}^{\mathrm{T}}\right\|_2^2
\end{equation}

Similarly, the loss from the coach to the student is:
\begin{equation}
\mathcal{L}^{\text{TG}}_{\text{C2S}} = \text{KL}\left( \frac{A^{\text{stu}}}{\tau} \bigg\| \frac{A^{\text{coa}}}{\tau} \right) + \lambda \cdot \left\|F_{\mathrm{BEV}}^{\mathrm{S}} - F_{\mathrm{BEV}}^{\mathrm{C}}\right\|_2^2
\end{equation}

The overall TGPD loss is a weighted combination of both supervision signals:
\begin{equation}
\mathcal{L}_{\text{TGPD}} = \beta_1 \cdot \mathcal{L}^{\text{TG}}_{\text{T2S}} + \beta_2 \cdot \mathcal{L}^{\text{TG}}_{\text{C2S}}
\end{equation}

\subsubsection{Masked Semantic Response Distillation (MSRD)}
\begin{figure}[H]
    \centering
    \includegraphics[width=0.5\textwidth]{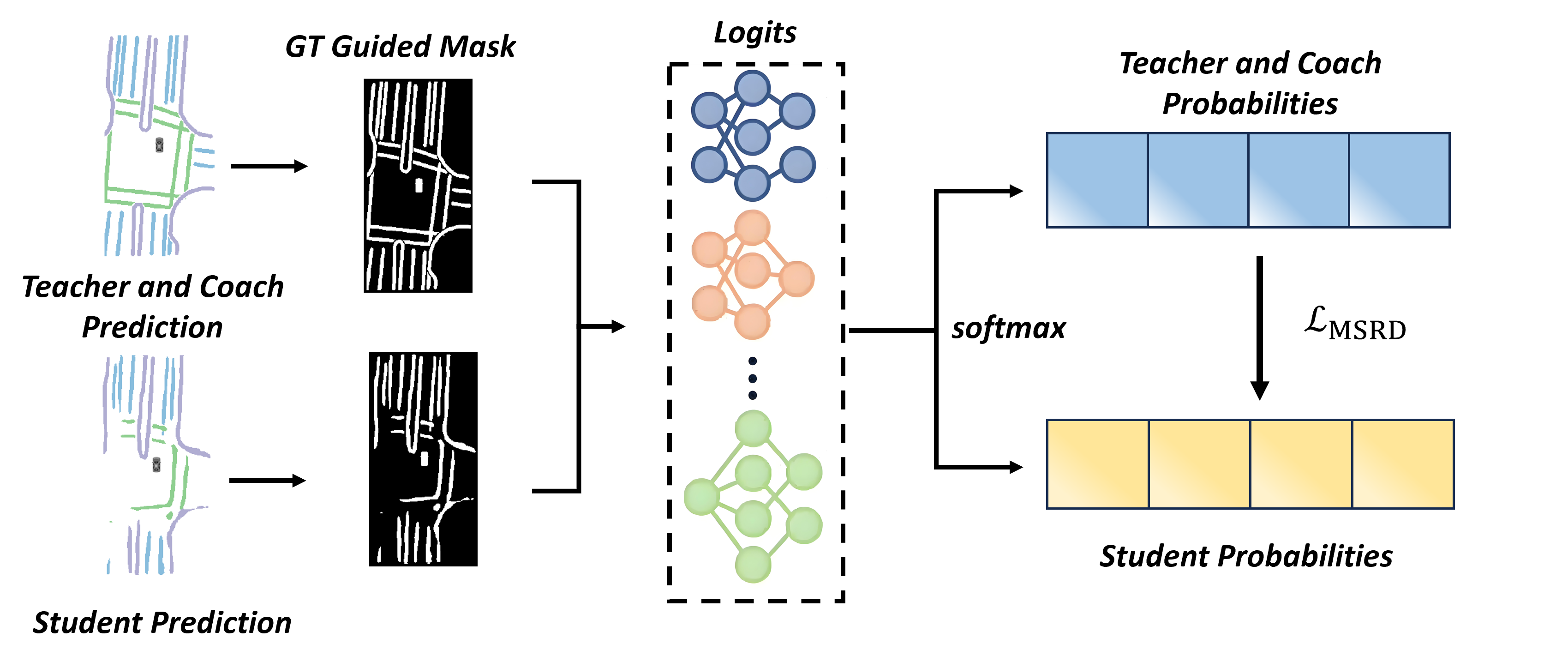}
    \caption{Illustration of the Masked Semantic Response Distillation (MSRD) process.}
    \label{fig:MSRD}
\end{figure}

.
MSRD distills semantic knowledge at the output level by aligning logits on valid foreground regions, which obtain map elements, as shown in Figure~\ref{fig:MSRD}. Let \( S_S, S_T, S_C \) represent the semantic logits from the student, teacher, and coach, and \( M \) be a binary mask derived from the ground truth as the ground truth-guided mask ensures distillation focuses on semantically meaningful regions, enhancing learning efficiency. We extract the masked logits:

\begin{equation}
\tilde{S} = S_S[M], \quad \tilde{T} = S_T[M], \quad \tilde{C} = S_C[M]
\end{equation}

We then compute the probabilities as follows:

\begin{equation}
P_T = \sigma(\tilde{T}), \quad P_C = \sigma(\tilde{C})
\end{equation}
The final loss is defined as:

\begin{equation}
\mathcal{L}_{\text{MSRD}} =\gamma_1 \cdot \text{BCE}(\tilde{S}, P_T) + \gamma_2 \cdot \text{BCE}(\tilde{S}, P_C)
\end{equation}
\noindent where \( \alpha \) is a balancing weight for the coach’s supervision.

This design ensures that only meaningful foreground predictions are aligned, reducing the influence of background noise.
\begin{table*}[t]
\centering
\scriptsize
\renewcommand{\arraystretch}{1.2}
\setlength{\tabcolsep}{3pt}

\begin{tabular}{lccccccccccccc}
\hline
\textbf{Method} & \textbf{Base/Stu. Mod} & \textbf{Coa. Mod} & \textbf{Tea. Mod} & \textbf{Epochs} 
& \textbf{IOU$_{ped.}$} & \textbf{IOU$_{div.}$} & \textbf{IOU$_{bou.}$} & \textbf{mIOU} 
& \textbf{AP$_{ped.}$} & \textbf{AP$_{div.}$} & \textbf{AP$_{bou.}$} & \textbf{mAP} & \textbf{FPS} \\
\hline
\textbf{HDMapNet} & \textbf{C} & \textbf{-} & \textbf{-} & \textbf{30} & \textbf{39.50} & \textbf{13.80} & \textbf{40.20} & \textbf{31.16} & \textbf{21.42} & \textbf{11.74} & \textbf{36.25} & \textbf{23.13} & \textbf{44.8} \\

\textbf{PMapNet$^{*}$} & \textbf{C \& sim-L \& SD \& HD} & \textbf{-} & \textbf{-} & \textbf{30} & \textbf{44.80} & \textbf{26.20} & \textbf{45.30} & \textbf{38.76} & \textbf{26.98} & \textbf{26.09} & \textbf{52.14} & \textbf{35.07} & \textbf{13.4} \\

\textbf{PMapNet} & \textbf{C \& L \& SD \& HD} & \textbf{-} & \textbf{-} & \textbf{30} & \textbf{56.60} & \textbf{41.70} & \textbf{64.80} & \textbf{54.36} & \textbf{39.16} & \textbf{34.37} & \textbf{62.84} & \textbf{45.46} & \textbf{10.6} \\

\hline
BEVDistill & C & - & C \& L$^{*}$ & 10 & 40.90 & 15.00 & 41.10 & 32.33\textsubscript{+1.17} & 21.11 & 15.00 & 38.86 & 24.99\textsubscript{+1.86} & 44.5 \\
UniDistill & C & - & C \& L$^{*}$ & 10 & 40.20 & 14.39 & 41.40 & 32.00\textsubscript{+0.84} & 22.88 & 14.39 & 39.06 & 25.44\textsubscript{+2.31} & 44.9 \\
MapDistill & C & - & C \& L$^{*}$ & 10 & 40.50 & 16.20 & 41.60 & 32.77\textsubscript{+1.61} & 22.44 & 15.86 & 42.39 & 26.90\textsubscript{+3.77} & 44.8 \\
BEVDistill & C & - & C \& L$^{*}$ & 30 & 42.80 & 17.93 & 43.50 & 34.74\textsubscript{+3.58} & 26.03 & 17.93 & 44.25 & 29.40\textsubscript{+6.27} & 44.8 \\
UniDistill & C & - & C \& L$^{*}$ & 30 & 43.10 & 19.75 & 43.30 & 35.38\textsubscript{+4.22} & 23.18 & 19.75 & 45.53 & 29.48\textsubscript{+6.35} & 44.9 \\
MapDistill & C & - & C \& L$^{*}$ & 30 & 41.90 & 20.11 & 43.70 & 35.23\textsubscript{+4.07} & 24.04 & 19.14 & 46.51 & 29.89\textsubscript{+6.76} & 44.8 \\
\underline{\textbf{MapKD (Ours)}} & \underline{\textbf{C}} & \underline{\textbf{C \& sim-L$^{*}$}} & \underline{\textbf{C \& L$^{*}$}} & \underline{\textbf{10}} & \underline{\textbf{44.40}} & \underline{\textbf{25.40}} & \underline{\textbf{43.71}} & \underline{\textbf{37.84\textsubscript{+6.68}}} & \underline{\textbf{27.30}} & \underline{\textbf{25.60}} & \underline{\textbf{49.32}} & \underline{\textbf{34.07\textsubscript{+10.94}}} & \underline{\textbf{44.9}} \\
\hline
\end{tabular}

\caption{\small  Comparison of our proposed MapKD with prior baselines and distillation methods under the camera-only setting. “Stu. Mod”, “Coa. Mod”, and “Tea. Mod” denote student, coach, and teacher input modalities. “C” and “L” refer to camera and LiDAR, "sim-L" indicates pseudo-LiDAR features. “C \& sim-L*” and "C \& L*" represent settings with SD and HD supervision. The first row represents the student model, the second row denotes the coach (PMapNet*), and the third row corresponds to our full-modality teacher. The underlined row highlights our proposed MapKD, which achieves the best performance with only 10 training epochs.}
\label{table:main_exp}
\end{table*}

\subsection{Training Strategy Design}
 In our TCS framework, the total training objective combines standard supervised loss with two types of distillation losses from both teacher and coach networks. The overall loss is formulated as:

\begin{equation}
\mathcal{L}_{\text{total}} = \mathcal{L}_{\text{base}} + \lambda_1 \mathcal{L}_{\text{bev}} + \lambda_2 \mathcal{L}_{\text{output}}
\end{equation}

 \( \mathcal{L}_{\text{base}} \) denotes the fundamental supervised loss guided by the ground truth, including semantic segmentation, instance embedding, and optional direction prediction:

\begin{equation}
\mathcal{L}_{\text{base}} = \mathcal{L}_{\text{seg}} + \alpha_1 \mathcal{L}_{\text{dist}} + \alpha_2 \mathcal{L}_{\text{dir}}
\end{equation}

 \( \mathcal{L}_{\text{bev}} \) represents bev-level distillation loss guided by the teacher and coach. It follows a cross-attention-based distillation (TGPD) strategy:

\begin{equation}
\mathcal{L}_{\text{bev}} = \beta_1 \cdot \mathcal{L}^{\text{TG}}_{\text{T2S}} + \beta_2 \cdot \mathcal{L}^{\text{TG}}_{\text{C2S}}
\end{equation}

 \( \mathcal{L}_{\text{output}} \) is the logit-level distillation loss that guides the student’s semantic output with probabilities from teacher and coach. It operates only on foreground regions as defined by a mask:

\begin{equation}
\mathcal{L}_{\text{output}} = \gamma_1 \cdot \text{BCE}(\tilde{S}, \sigma(\tilde{T})) + \gamma_2 \cdot \text{BCE}(\tilde{S}, \sigma(\tilde{C}))
\end{equation}

\noindent where \( \tilde{S}, \tilde{T}, \tilde{C} \) denote the student, teacher, and coach semantic logits respectively, and \( \sigma(\cdot) \) is the sigmoid activation function.

\vspace{-3mm}
\section{Experiments}

\subsection*{Experimental setups}

\subsubsection{Dataset.} We conduct our experiments on the nuScenes dataset\cite{caesar2020nuscenes}, a large-scale autonomous driving benchmark containing multimodal sensor data collected in urban scenes. Each sample contains multi-view images, ego-pose, and corresponding semantic labels . Following the settings in recent work \cite{jiang2024pmapnet}, we select three types of map elements for fair comparison: pedestrian crossings, lane dividers, and road boundaries.

\subsection*{Implementation Details}
In the first stage, the teacher and coach model is jointly pretrained.

In the second stage, the student model  is trained from scratch under the proposed TCS pipeline. The teacher and the coach are kept frozen, while  providing intermediate guidance at the feature and output level. The student is supervised with both hard labels and soft logits.All training hyperparameters and implementation details are provided in Appendix C.

\subsection*{Experiment Results}
\subsubsection{Comparison with Baselines and Existing Distillation Methods.}
The comparison between MapKD and other state-of-the-art  methods is presented in Table \ref{table:main_exp}. The pre-training results of the student, coach and teacher models used in MapKD are reported in Rows 1, 2 and 3 of the table, respectively. As mentioned in the Methodology section, the lightweight, vision-centric HDMapNet is adopted as the student model, while the PMapNet model incorporating SD map and HD map priors is employed for both the coach and teacher branches. The coach model takes only surround-view images as input and further transforms them into image features and pseudo-LiDAR features, in contrast to the teacher model, which processes multi-modal data.
In the second part of the table, we present a comparison between MapKD and other advanced distillation approaches adopting the conventional two-stage Teacher–Student paradigm, BEVDistill, UniDistill and MapDistill. For a fair comparison, these baseline distillation methods use the same teacher model (PMapNet with multi-modal inputs) and the same student model (vision-centric HDMapNet) as MapKD. Regarding training iterations, the baseline distillation methods are trained for 10 epochs and 30 epochs, respectively.
\begin{figure*}[!t]
    \centering
    \includegraphics[width=\textwidth]{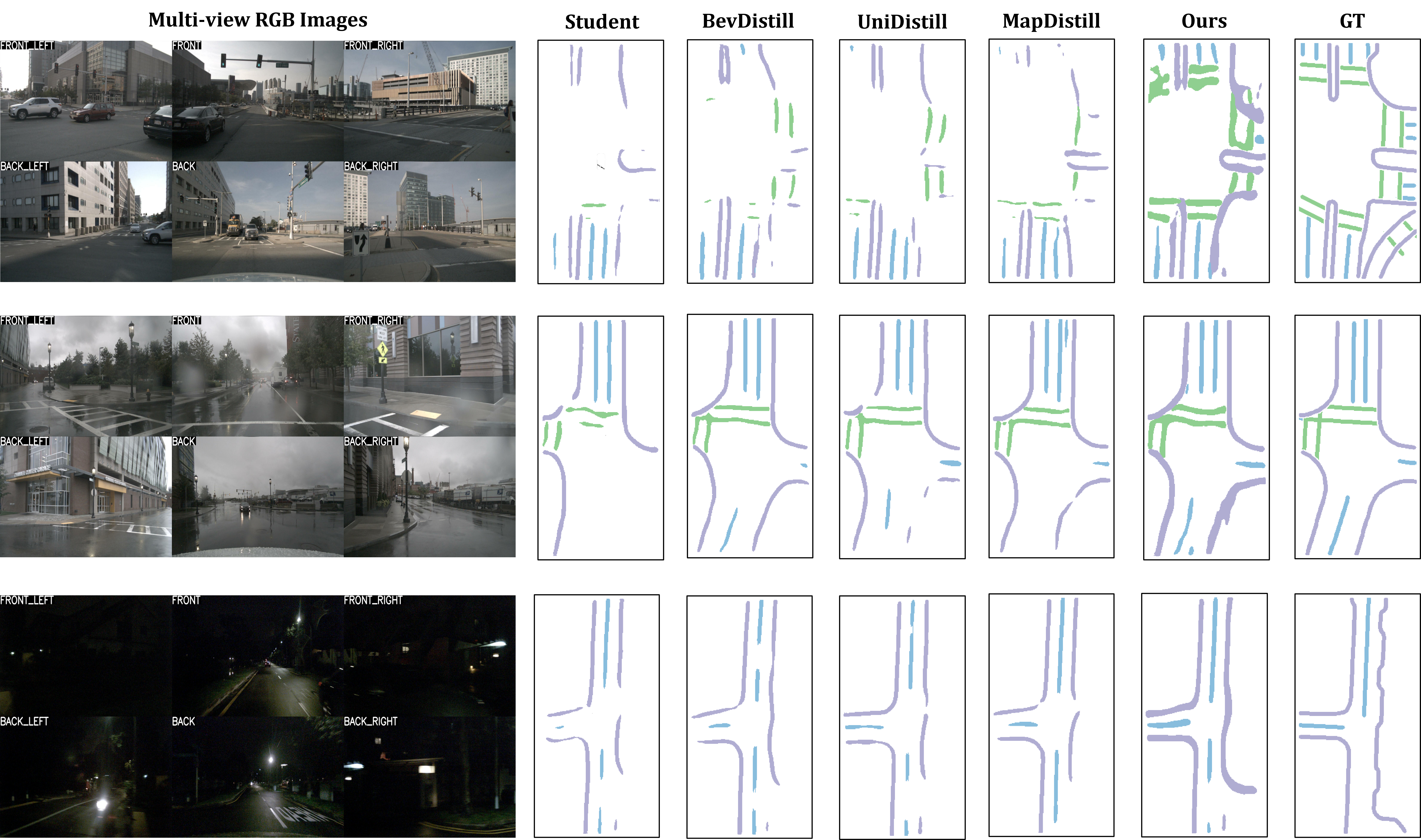}
    \caption{Qualitative comparison of HD map generation under different scenarios. Our MapKD outperforms the student model without distillation (HDMapNet) and MapDistill, providing coherent and complete maps with accurate details.}
    \label{fig:vis1}
\end{figure*}

Table \ref{table:main_exp} shows that:
(1) Comparing Rows 1, 2, 3 with the final row, under the joint effect of the TCS distillation framework and the TGPD and MSRD distillation loss functions, MapKD achieves 37.84 mIoU and 34.07 mAP without relying on SD map or HD map priors, using only camera data as input. This performance surpasses the baseline method HDMapNet (with the same input modality) by +6.68 mIoU and +10.94 mAP. This indicates that MapKD effectively transfers the geometric information and prior knowledge embedded in the teacher and coach branches to the student model in an implicit manner, enabling efficient knowledge transfer.
(2) While traditional distillation methods do improve performance over the baseline student model, their results under the same 10 epochs training schedule remain inferior to MapKD.
(3) Compared with other advanced distillation methods (as shown in Rows 4, 5, and 6), MapKD surpasses the performance of these methods trained for 30 epochs while requiring only 10 epochs. This demonstrates that the synergy of the TCS distillation framework with TGPD and MSRD loss functions enables faster and more effective transfer of knowledge from large models to the student model.Qualitative visualization results in Figure \ref{fig:vis1} further evidence MapKD’s superiority.
\begin{table}[htbp]

\centering
\scriptsize
\renewcommand{\arraystretch}{1.2}
\setlength{\tabcolsep}{5pt}
\begin{tabular}{lccc|cc}
\hline
\textbf{Method} & \textbf{Modality} & \textbf{Epochs} & \textbf{Distilled} & \textbf{mIOU} & \textbf{mAP} \\
\hline
Student                   & C               & 30 & \ding{55} & 31.16 & 23.13 \\
Student with Priors       & C + (SD)        & 30 & \ding{55} & 37.01 & 27.84 \\
Student with Priors       & C + (SD + HD)   & 30 & \ding{55} & \textbf{38.10} & \underline{33.63} \\
\textbf{MapKD (Ours)}     & \textbf{C}      & \textbf{10} & \textbf{\ding{51}} & \underline{37.84} & \textbf{34.07} \\
\hline
\end{tabular}
\caption{\small Comparison of training settings with or without map priors and distillation supervision.}
\label{table:t2}
\end{table}

\subsubsection{Eliminating the Dependency on Map Priors.}
In Table \ref{table:t2}, it shows that compared with the baseline student model, incorporating SD map priors improves mIoU, and further adding HD map pre-training increases mAP. However, with our proposed MapKD distillation method, the model—despite using only surround-view image inputs—still outperforms the variant that relies on both SD map and HD map dependencies. These results highlight that MapKD can effectively learn structural priors without relying on costly map priors or their annotations.

\subsection{Ablation Studies.}
\subsubsection{Ablation on Distillation Architecture.}

To assess the effectiveness of each component in our proposed distillation framework, we conduct an ablation study focusing on three key losses in Table~\ref{table:3}: (1) the TCS supervision loss ($\mathcal{L}_{\text{TCS}}$), which enables hierarchical knowledge transfer through the intermediate coach; (2) the Token-Guided Patch Distillation loss ($\mathcal{L}_{\text{TGPD}}$), which aligns BEV features across modalities; and (3) the Masked Semantic Response Distillation loss ($\mathcal{L}_{\text{MSRD}}$), which supervises segmentation predictions.

\begin{table}[htbp]
\centering
\scriptsize
\renewcommand{\arraystretch}{1.2}
\setlength{\tabcolsep}{10pt}  
\begin{tabular}{lccccc}
\hline
\textbf{Range} & \textbf{$\mathcal{L}_{\text{TCS}}$} & \textbf{$\mathcal{L}_{\text{TGPD}}$} & \textbf{$\mathcal{L}_{\text{MSRD}}$} & \textbf{mIOU} & \textbf{mAP} \\
\hline
Baseline & \ding{55} & \ding{55} & \ding{55} & 31.16 & 23.13 \\
a        & \ding{55} & \ding{55} & \ding{51} & 33.30 & 29.64 \\
b        & \ding{55} & \ding{51} & \ding{55} & 36.00 & 31.36 \\
c        & \ding{55} & \ding{51} & \ding{51} & 37.40 & 31.90 \\
d        & \ding{51} & \ding{55} & \ding{51} & 37.30 & 32.87 \\
e        & \ding{51} & \ding{51} & \ding{55} & 37.73 & 32.29 \\
\textbf{f} & \textbf{\ding{51}} & \textbf{\ding{51}} & \textbf{\ding{51}} & \textbf{37.84} & \textbf{34.07} \\
\hline
\end{tabular}
\caption{\small Ablation study on the components of our MapKD distillation method. }
\label{table:3}
\end{table}

In Rows a, b, and c, we evaluate the effectiveness of the two distillation strategies—TGPD and MSRD—by applying each individually to guide the student model in learning knowledge from the teacher model. The results show that both strategies bring notable improvements over the baseline student model, and their combination further enhances the distillation performance, validating the effectiveness of both strategies.
In Rows d and e, the above distillation strategies are evaluated within the TCS framework. The results indicate that introducing the coach model, even when paired with only a single distillation strategy, yields additional improvements in either semantic transfer or feature alignment. This demonstrates that the TCS framework can enhance the efficiency of knowledge transfer across different representation levels.
In Row f, the TCS framework is combined with both distillation strategies, achieving the best performance (37.84 mIoU / 34.07 mAP), representing improvements of +6.68 mIoU and +10.94 mAP over the baseline model. These results further confirm the advantage of jointly applying the TCS architecture and the two distillation strategies for cross-modal knowledge transfer.

\begin{table}[htbp]
\centering
\scriptsize
\renewcommand{\arraystretch}{1.2}
\setlength{\tabcolsep}{3pt}
\begin{tabular}{lccc|cc}
\hline
\textbf{Framework} & \textbf{Teacher} & \textbf{Coach} & \textbf{Student} & \textbf{mIoU} & \textbf{mAP} \\
\hline
Baseline & \textbackslash
 & \textbackslash
 & HDMapNet & 31.16 & 23.13 \\
Two-stage & HDMapNet (L+C) & \textbackslash
 & HDMapNet & 32.58 & 25.96 \\
Two-stage & PMapNet & \textbackslash
 & HDMapNet & 37.40 & 31.91 \\
Two-stage & PMapNet* & \textbackslash
 & HDMapNet & 33.13 & 27.36 \\
\textbf{Ours (Three-stage)} & \textbf{PMapNet} & \textbf{PMapNet*} & \textbf{HDMapNet} & \textbf{37.84} & \textbf{34.07} \\
\hline
\end{tabular}
\caption{\small Comparison of two-stage and three-stage distillation frameworks. The PMapNet* is our coach model.}
\label{tab:4}
\end{table}

\subsubsection{Ablation on Two-Stage vs Three-Stage Distillation Framework.}
We conduct an ablation study comparing traditional two-stage distillation with our proposed three-stage TCS framework. As shown in Table~\ref{tab:4}, even when using the best teacher (PMapNet) with our TGPD and MSRD losses, the two-stage setting achieves only 37.40 mIoU and 31.91 mAP. In contrast, our three-stage approach reaches 37.84 mIoU and 34.07 mAP, demonstrating superior effectiveness. The results confirm that introducing a coach network significantly enhances knowledge transfer by bridging the modality gap between the teacher and student.

\begin{table}[htbp]
\centering
\scriptsize
\renewcommand{\arraystretch}{1.2}
\setlength{\tabcolsep}{10pt}
\begin{tabular}{lcc|cc}
\hline
\textbf{Settings} & $L_{\text{bev}}$ & $L_{\text{output}}$ & \textbf{mIoU} & \textbf{mAP} \\
\hline
Three-stage \& BEVDistill & \ding{51} & \ding{55} & 37.54 & 31.70 \\
Three-stage \& UniDistill & \ding{51} & \ding{55} & 37.43 & 31.65 \\
Three-stage \& MapDistill & \ding{51} & \ding{51} & 37.70 & 33.12 \\
\textbf{Ours (MapKD)} & \textbf{\ding{51}} & \textbf{\ding{51}} & \textbf{37.84} & \textbf{34.07} \\
\hline
\end{tabular}
\caption{\small Comparison of distillation strategies under the three-stage framework. Our method achieves the best performance.}
\label{tab:5}
\end{table}

\subsubsection{Ablation on Distillation Strategies within the TCS Framework.}
In this ablation study, we compare different distillation strategies within the unified three-stage TCS framework. Specifically, we replace our default loss functions with BEVDistill~\cite{chen2023bevdistill} (which aligns BEV features via MSE), UniDistill~\cite{zhou2023unidistill} (which performs token-level attention alignment) and MapDistill~\cite{hao2024mapdistill} (which combines feature and output-level supervision), while keeping all other settings fixed.
As shown in Table~\ref{tab:5}, our proposed TGPD+MSRD achieves the best performance, reaching 37.84 mIoU and 34.07 mAP. These results highlight the superior effectiveness and compatibility of our distillation method within the TCS framework.

\subsubsection{Ablation on Distillation Loss Weights and Impact.}
\begin{figure}[t]
    \centering
    \includegraphics[width=0.9\linewidth]{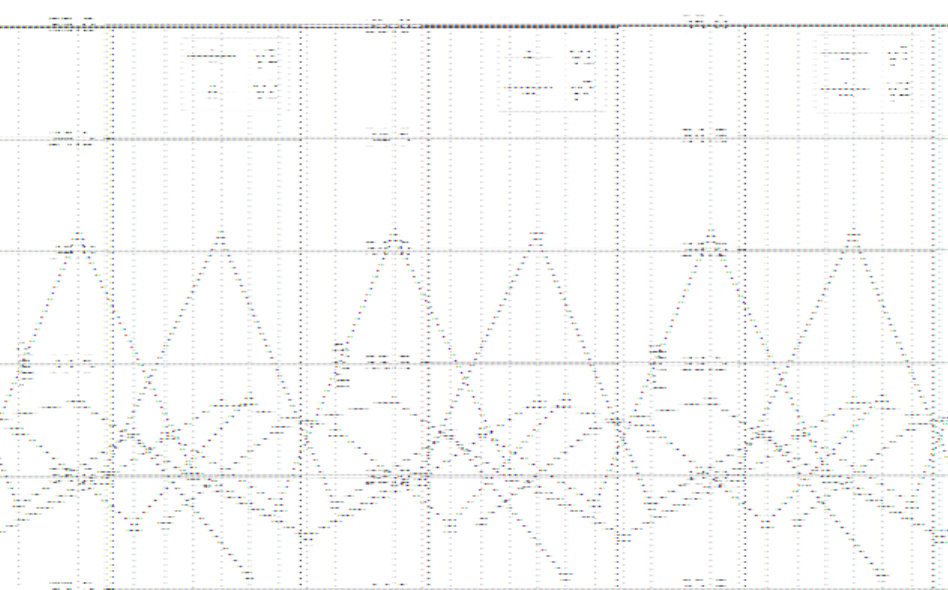}
    \caption{Performance of different TGPD loss weights.}
    \label{fig:loss1}
\end{figure}

\begin{figure}[t]
    \centering
    \includegraphics[width=0.9\linewidth]{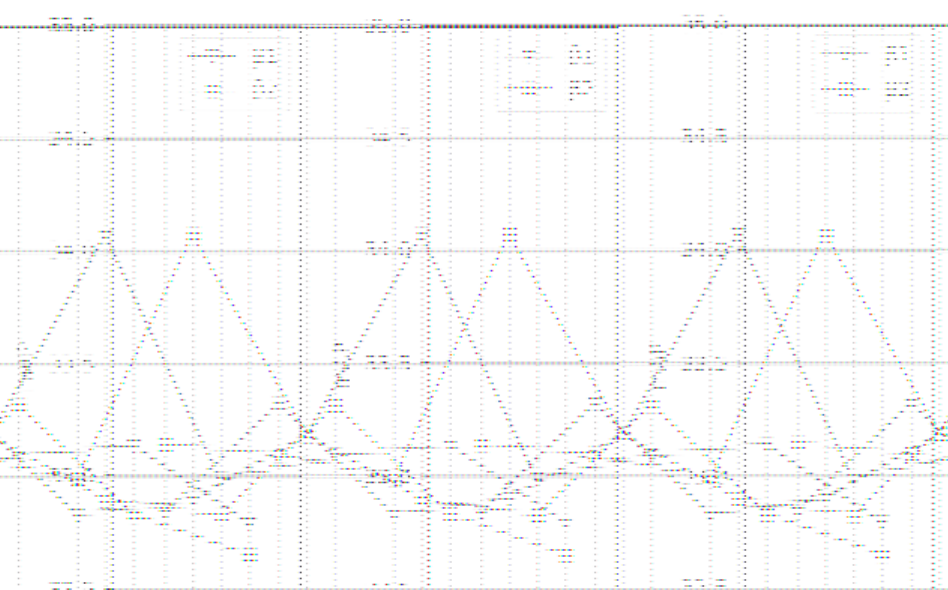}
    \caption{Performance of different MSRD loss weights.}
    \label{fig:loss2}
\end{figure}

In this experiment, we investigate the impact of loss weighting parameters in our three-stage distillation framework. Specifically, we vary the values of $\beta_1$ and $\beta_2$, which control the contributions of TGPD loss from the teacher and coach respectively, as well as $\gamma_1$ and $\gamma_2$, which control the MSRD loss contributions.  For each parameter under study, we vary it within a reasonable range while keeping the other parameters fixed. As shown in Figure~\ref{fig:loss1} and Figure~\ref{fig:loss2}, we observe that the best performance is achieved when $\beta_1 = 0.6$, $\beta_2 = 0.4$, $\gamma_1 = 0.7$ and $\gamma_2 = 0.3$. Under this configuration, our method achieves the highest accuracy, demonstrating the importance of balanced supervision from both teacher and coach in optimizing student learning.

\vspace{-2mm}
\section{Conclusion}

In this paper, we proposed \textbf{MapKD}, a novel three-stage teacher–coach–student distillation framework for camera-centric HD map construction.  The coach uses simulated LiDAR to bridge the gap between camera and LiDAR inputs, allowing the student to learn from the teacher without using HD map labels. We also design two distillation strategies: Token-Guided Patch Distillation (TGPD) for supervising BEV features, and Masked Semantic Response Distillation (MSRD) for guiding semantic predictions. Experiments on nuScenes dataset show that MapKD performs well both quantitatively and qualitatively, offering a flexible and efficient solution without relying on HD map pretraining.

\clearpage
\section*{Appendix}
\subsection*{Appendix A: Motivation for Designing the Coach in TCS Distillation}
In MapKD framework, the student model receives only camera images as input, as shown in Algorithm~\ref{alg:student}. In contrast, the teacher model uses full-modal inputs, including LiDAR,SD maps and HD maps, as shown in Algorithm~\ref{alg:teacher}. This causes clear gaps between the two models. The gaps can be described from three perspectives:

(1): \textbf{modality gap}. The teacher observes the 3D world through LiDAR,SD maps and HD maps. These inputs offer precise geometry and reliable structure. The student observes only raw RGB images, which are often affected by lighting, occlusion, or perspective changes. As a result, the teacher and student focus on different scales of the scene and extract different types of features.

\begin{figure*}[!t]
    \centering
    \includegraphics[width=\linewidth]{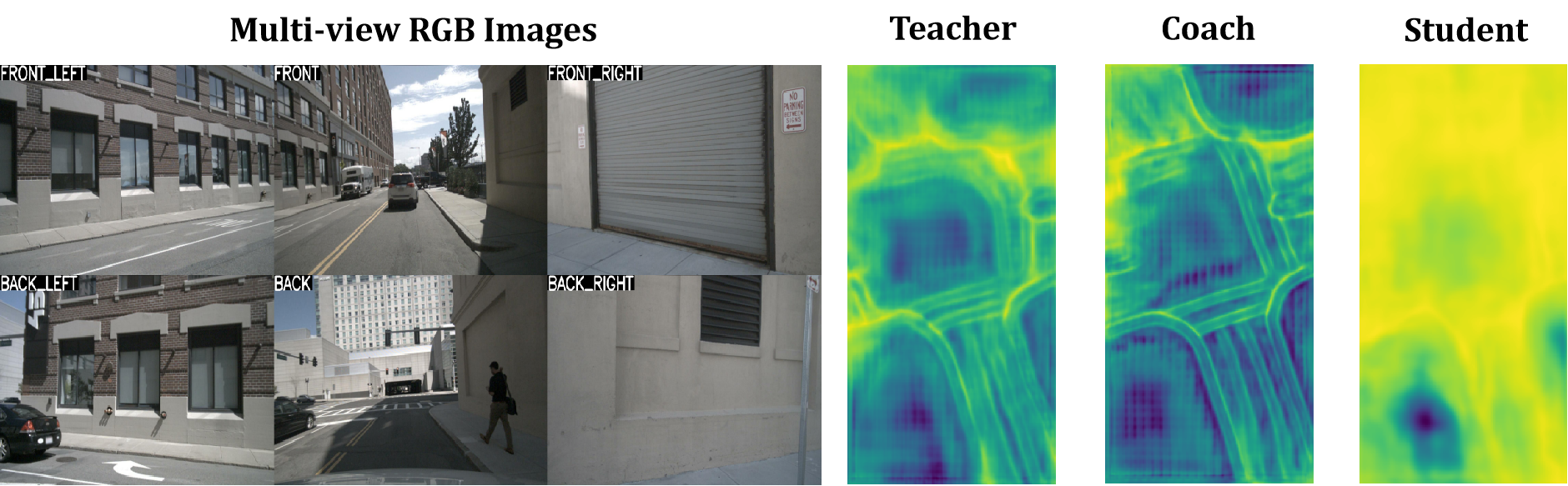}
     \caption{
    BEV feature comparison among teacher, coach, and student(w/o MapKD).
The student shows blurry results; the coach improves structure; the teacher yields the most detailed features.  
  }
    \label{fig:bevtcs}
\end{figure*}

(2): \textbf{representation gap}. As illustrated in Figure~\ref{fig:bevtcs}, the BEV features generated by the teacher are rich in 3D-aware patterns. They often highlight drivable areas, object boundaries, and elevation cues. The student, however, struggles to capture such structured signals from monocular views. Its BEV features are typically blurry, fragmented, or misaligned. When the teacher tries to guide the student directly via two-stage paradigm, the student may not be able to interpret or absorb the teacher’s knowledge well.

(3): \textbf{learning gap}. Despite receiving strong and detailed supervision from the teacher, the student model often fails to effectively learn or internalize this knowledge. The gap arises not only from input limitations, but also from the student’s insufficient capacity to interpret spatial cues and abstract structured patterns. This leads to suboptimal feature learning and reduced generalization capability.

To address all of these gaps, we introduce the \textbf{coach model}. The coach uses only camera images, just like the student, as shown in Algorithm~\ref{alg:coach}. However, it is equipped with an extra module that generates pseudo-LiDAR features. These features imitate geometric cues found in real LiDAR data but are inferred from images. This design gives the coach two key advantages.

First, the coach operates in a \textbf{modality-aligned space}. It receives the same visual input as the student and learns to extract similar information. This makes its internal features easier for the student to follow. In other words, the student can learn from the coach more naturally, without struggling to understand LiDAR-based features.

Second, the coach serves as a \textbf{representation translator}. It learns from the teacher how to organize its BEV features into structured maps. Then, it passes this structure to the student in a more interpretable form. This makes the distillation process smoother and more effective. We summarize the overall performance comparison between our proposed method and traditional baselines in \textbf{Table~1}, and further provide a detailed comparison between two-stage and three-stage distillation strategies in \textbf{Table~\ref{tab:ablation_hdmap_usage}}. These results consistently validate the effectiveness of our proposed framework.

\begin{algorithm}[tb]
\caption{Student Model Inference Process}
\label{alg:student}
\textbf{Input}: Multi-view images $I$ \\
\textbf{Output}:Student BEV feature $F_{BEV}^{\mathrm{S}}$, Final semantic map $S_{\mathcal{S}}$
\begin{algorithmic}[1]
\STATE Extract image features: $F_I^{\mathcal{S}} \gets \text{Enc}_{\mathrm{img}}(I)$
\STATE Project to BEV: $F_{BEV}^{\mathrm{S}} \gets \text{BEVProj}(F_I^{\mathcal{S}})$
\STATE Decode semantics: $S_{\mathcal{S}} \gets \text{Dec}_{\mathrm{sem}}(F_{BEV}^{\mathrm{S}})$
\STATE \textbf{return} $F_{BEV}^{\mathrm{S}},S_{\mathcal{S}}$
\end{algorithmic}
\end{algorithm}

\begin{algorithm}[tb]
\caption{Teacher Model Pipeline}
\label{alg:teacher}
\textbf{Input}: Multi-view images $I$, LiDAR $L$, Pretraining HD map $M_{HD}^{\text{noisy}}$, HD map $M_{HD}$ \\
\textbf{Output}: Teacher BEV feature $F_{BEV}^{\mathrm{T}}$, Semantic prediction $S_{\mathcal{T}}$
\begin{algorithmic}[1]
\STATE Extract image features: $F_I^{\mathcal{T}} \gets \text{Enc}_{\mathrm{img}}(I)$
\STATE Encode LiDAR features: $F_L \gets \text{Enc}_{\mathrm{LiDAR}}(L)$
\STATE Project to BEV: $F_{\mathrm{BEV}} \gets \text{BEVProj}(F_I^{\mathcal{T}}, F_L)$
\STATE Fuse with SD map: $F_{\mathrm{fused}} \gets \text{SDFusion}(F_{BEV}^{\mathrm{T}}, M_{SD})$
\STATE \textbf{[Pretrain]} Recover structure from HD map: $S_{\mathcal{T}} \gets \text{Dec}_{\mathrm{sem}}(\text{HDFusion}(F_{\mathrm{fused}}, M_{HD}^{\text{noisy}}))$
\STATE \textbf{return} $F_{BEV}^{\mathrm{T}},S_{\mathcal{T}}$

\end{algorithmic}
\end{algorithm}

\begin{algorithm}[tb]
\caption{Coach Model Pipeline}
\label{alg:coach}
\textbf{Input}: Multi-view images $I$, SD map $M_{SD}$, Pretraining HD map $M_{HD}^{\text{noisy}}$, HD map $M_{HD}$ \\
\textbf{Output}:Coach BEV feature $F_{BEV}^{\mathrm{C}}$, Semantic map $S_{\mathcal{C}}$
\begin{algorithmic}[1]
\STATE Extract image features: $F_I^{\mathcal{C}} \gets \text{Enc}_{\mathrm{img}}(I)$
\STATE Generate pseudo-LiDAR: $\tilde{F}_L \gets \text{PseudoLiDAR}(I)$
\STATE Project to BEV: $F_{\mathrm{BEV}} \gets \text{BEVProj}(F_I^{\mathcal{C}}, \tilde{F}_L)$
\STATE Fuse with SD map: $F_{\mathrm{fused}} \gets \text{SDFusion}(F_{BEV}^{\mathrm{T}}, M_{SD})$
\STATE \textbf{[Pretrain]} Recover structure from HD map: $S_{\mathcal{C}} \gets \text{Dec}_{\mathrm{sem}}(\text{HDFusion}(F_{\mathrm{fused}}, M_{HD}^{\text{noisy}}))$

\STATE \textbf{return} $F_{BEV}^{\mathrm{C}},S_{\mathcal{C}}$

\end{algorithmic}
\end{algorithm}

\begin{table*}[t]
\centering
\scriptsize
\renewcommand{\arraystretch}{1.2}
\setlength{\tabcolsep}{5pt}  
\begin{tabular}{lccccccccccccc}
\hline
\textbf{Method} & \textbf{Mod} & \textbf{Role} & \textbf{Epochs} 
& \textbf{IOU$_{ped.}$} & \textbf{IOU$_{div.}$} & \textbf{IOU$_{bou.}$} & \textbf{mIOU} 
& \textbf{AP$_{ped.}$} & \textbf{AP$_{div.}$} & \textbf{AP$_{bou.}$} & \textbf{mAP} & \textbf{FPS} \\
\hline   
\underline{\textbf{HDMapNet}} & \underline{\textbf{C}} & \underline{\textbf{student}} & \underline{\textbf{30}} & \underline{\textbf{39.50}} & \underline{\textbf{13.80}} & \underline{\textbf{40.20}} & \underline{\textbf{31.16}} & \underline{\textbf{21.42}} & \underline{\textbf{11.74}} & \underline{\textbf{36.25}} & \underline{\textbf{23.13}} & \underline{\textbf{44.8}} \\
HDMapNet & C \& L & - & 30 & 45.80 & 30.51 & 56.83 & 44.40 & 29.46 &  13.89 & 54.07 &  32.47 & 20.4 \\
PMapNet & C \& SD & - & 30 & 44.04 & 23.70 & 43.30 & 37.01 & 22.51 & 14.39 & 46.62 & 27.84 & 32.2 \\
PMapNet & C \& SD \& HD & - & 30 & 45.03 & 24.61 & 43.34 & 37.66 & 28.07 & 23.93 & 48.91 & 33.63 & 14.3 \\
PMapNet & C \& L\& SD & - & 30 & 55.29 & 40.65 & 63.75 & 53.23 & 28.21 & 26.40 & 54.37 & 36.32 & 20.2 \\
\underline{\textbf{PMapNet}} & \underline{\textbf{C \& L\& SD \& HD}} & \underline{\textbf{teacher}} & \underline{\textbf{30}} & \underline{\textbf{56.60}} & \underline{\textbf{41.70}} & \underline{\textbf{64.80}} & \underline{\textbf{54.36}} & \underline{\textbf{39.16}} & \underline{\textbf{34.37}} & \underline{\textbf{62.84}} & \underline{\textbf{45.46}} & \underline{\textbf{10.6}} \\
PMapNet$^{*}$ & C \& sim-L \& SD & - & 30 & 44.26 & 25.08 & 44.73 & 38.02 & 25.91 & 24.55 & 52.04 & 34.16 & 22.9 \\

\underline{\textbf{PMapNet$^{*}$}} & \underline{\textbf{C \& sim-L \& SD \& HD}} & \underline{\textbf{coach}} & \underline{\textbf{30}} & \underline{\textbf{44.80}} & \underline{\textbf{26.20}} & \underline{\textbf{45.30}} & \underline{\textbf{38.76}} & \underline{\textbf{26.98}} & \underline{\textbf{26.09}} & \underline{\textbf{52.14}} & \underline{\textbf{35.07}} & \underline{\textbf{13.4}} \\
\hline
\underline{\textbf{MapKD (Ours)}} & \underline{\textbf{C}} & \underline{\textbf{student}} & \underline{\textbf{10}} & \underline{\textbf{44.40}} & \underline{\textbf{25.40}} & \underline{\textbf{43.71}} & \underline{\textbf{37.84\textsubscript{+6.68}}} & \underline{\textbf{27.30}} & \underline{\textbf{25.60}} & \underline{\textbf{49.32}} & \underline{\textbf{34.07\textsubscript{+10.94}}} & \underline{\textbf{44.9}} \\
\hline
\end{tabular}

\caption{\small  Comparison of our proposed MapKD with prior baselines and distillation methods under the camera-only setting. “Stu. Mod”, “Coa. Mod”, and “Tea. Mod” denote student, coach, and teacher input modalities. “C” and “L” refer to camera and LiDAR, \textit{sim-L} indicates pseudo-LiDAR features. “SD \& HD” represent settings with SD and HD supervision. The first underlined row represents the student baseline, the second underlined row corresponds to our full-modality teacher, and the third underlined row denotes the coach. The fourth underlined row highlights our proposed MapKD, which achieves the best performance with only 10 training epochs.}
\end{table*}
\subsection*{Appendix B: Related Work}
Light-weight Online mapping is vital for autonomous driving, aiming to build semantic maps in dynamic environments. In recent years, numerous methods for online HD map generation have emerged. In this paper, We analyze these approaches along three main dimensions.
\subsubsection{Online Semanticize HD mapping.}

Traditional SLAM methods have been widely applied in autonomous driving, but most of them are constructed offline, making it difficult to meet real-time requirements and lacking rich semantic information. To address these limitations, researchers have started exploring online semantic high-definition (HD) map construction methods, aiming to generate semantically enriched HD maps in real-time using onboard sensors.
Early online mapping approaches primarily relied on image-based depth estimation and projection techniques, such as Lift-Splat-Shoot (LSS)~\cite{philion2020lss}, PON~\cite{yang2021projecting}, and VPN~\cite{hu2021vpn}. These methods generate bird’s-eye-view semantic representations from single or multiple frames, but they still suffer from limitations in geometric accuracy and semantic consistency. Later, HDMapNet~\cite{li2022hdmapnet} was proposed as a representative advancement in multi-sensor fusion. It utilizes both camera and LiDAR inputs to perform dense semantic prediction, followed by geometric post-processing to extract map elements such as lanes and road boundaries.
Despite these advances, online semantic HD map construction still faces several challenges. For instance, the dependence on high-precision sensors such as LiDAR increases system cost. Moreover, in large-scale HD map construction scenarios, the strong reliance on sensor observations often leads to reduced accuracy, limiting their effectiveness in real-world applications.
\begin{table}[H]
\centering
\scriptsize
\renewcommand{\arraystretch}{1.2}
\setlength{\tabcolsep}{1.5pt} 
\begin{tabular}{lccc|c}
\hline
\textbf{Framework} & \textbf{T} & \textbf{C} & \textbf{S} & \textbf{mAP} \\
\hline
Baseline & -- & -- & HDMapNet (C) & 23.13 \\
Two-stage & HDMapNet (L+C) & -- & HDMapNet (C) & 28.19 \\
Two-stage & PMapNet (C+SD) & -- & HDMapNet (C) & 26.95 \\
Two-stage & PMapNet (C+SD+HD) & -- & HDMapNet (C) & 27.21 \\
Two-stage & PMapNet (L+C+SD+HD) & -- & HDMapNet (C) & 31.91 \\
Two-stage & PMapNet* & -- & HDMapNet (C) & 27.36 \\
\hline
\underline{Three-stage} & \underline{PMapNet (L+C+SD)} & \underline{PMapNet* (SD)} & \underline{HDMapNet (C)} & \underline{29.15} \\
\underline{Three-stage} & \underline{PMapNet (L+C+SD+HD)} & \underline{PMapNet* (SD+HD)} & \underline{HDMapNet (C)} & \underline{\textbf{34.07}} \\
\hline
\end{tabular}
\caption{\small Comparison of two-stage, and our three-stage distillation under different training configurations.}
\label{tab:ablation_hdmap_usage}
\end{table}
 
\subsubsection{HD Map Enhancement.}

To overcome the limitations of traditional online semantic HD map construction—such as low accuracy in large-scale mapping scenarios, recent works~\cite{xiong2023neural,jiang2024pmapnet,priorMap2023,gao2023satellite,liu2021realtime,yan2023superfusion,xia2023mvmap,li2023hdmapgen,kong2023mapvr} have explored incorporating offline map priors to enhance mapping accuracy under complex conditions.
NeuralMapPrior (NMP)~\cite{xiong2023neural} improves robustness by sharing map tiles generated by other vehicles under favorable weather conditions, enabling better predictions in adverse environments through cross-vehicle prior reuse.
PriorMapNet~\cite{priorMap2023} injects prior BEV queries and reference points into a hierarchical Transformer decoder, enhancing the detection of fine-grained road elements such as road boundaries and crosswalks.
PMapNet~\cite{jiang2024pmapnet} leverages both static navigation maps (SDMap) and HD map priors. It first produces a coarse map and subsequently refines it using a pretrained HD map prior distribution, thereby improving the spatial accuracy of the final output.
Although these methods enhance performance in challenging conditions, they typically depend on expensive offline map assets and introduce complex model architectures, which hinder real-time scalability and deployment.

\begin{algorithm}[tb]
\caption{TCS Distillation Framework Training}
\label{alg:tcs_training}
\textbf{Input:} Multi-view image $I$, LiDAR $L$, maps $M_{\mathrm{SD}}, M_{\mathrm{HD}}$, ground truth labels $Y$ \\
\textbf{Initialize:} Teacher $\mathcal{T}$, Coach $\mathcal{C}$, Student $\mathcal{S}$; weights $\alpha_1, \alpha_2, \beta_1, \beta_2$
\begin{algorithmic}[1]
\STATE // Teacher forward
\STATE \textbf{return} $F_{\mathcal{T}}$,$S_{\mathcal{T}}$

\STATE // Coach forward
\STATE \textbf{return} $F_{\mathcal{C}}$,$S_{\mathcal{C}}$

\STATE // Student forward
\STATE \textbf{return} $F_{\mathcal{S}}$,$S_{\mathcal{S}}$

\STATE // Compute loss
\STATE $\mathcal{L}_{\text{base}} \gets \mathcal{L}_{\text{seg}} + \alpha_1 \mathcal{L}_{\text{dist}} + \alpha_2 \mathcal{L}_{\text{dir}}$
\STATE $\mathcal{L}_{\text{feat}} \gets \text{TGPD}(F_{\mathrm{S}}, F_{\mathrm{T}}, F_{\mathrm{C}})$
\STATE $\mathcal{L}_{\text{logit}} \gets \text{MSRD}(S_S, S_T, S_C)$
\STATE $\mathcal{L}_{\text{total}} \gets \mathcal{L}_{\text{base}} + \beta_1\mathcal{L}_{\text{bev}} + \beta_2\mathcal{L}_{\text{output}}$

\STATE // Backpropagation
\STATE Update $\mathcal{S}$ with $\mathcal{L}_{\text{total}}$
\end{algorithmic}
\end{algorithm}
\begin{figure*}[!t]
    \centering
    \includegraphics[width=\linewidth]{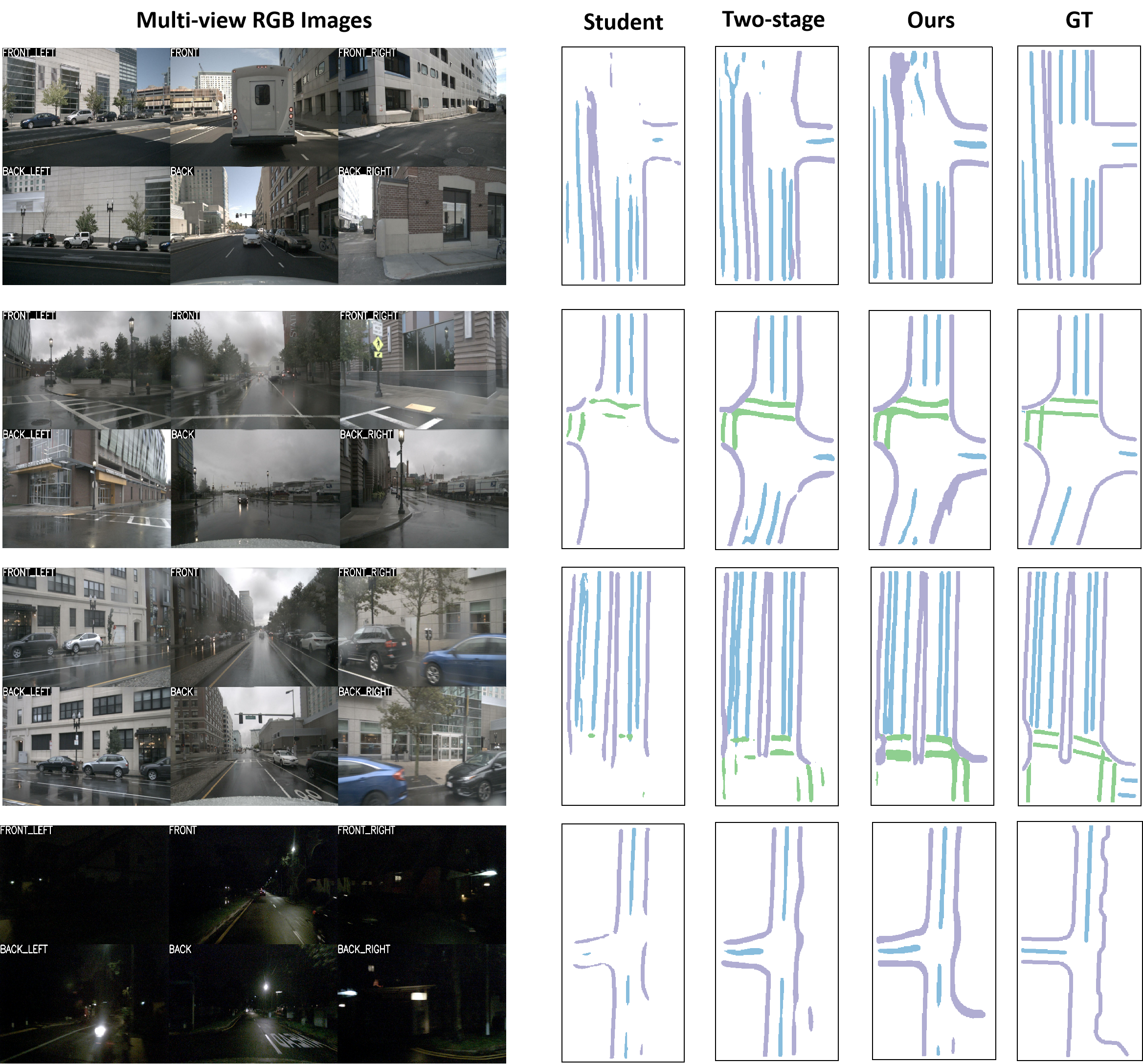}
    \caption{
    \textbf{Multi-scene evaluation of two-stage vs. three-stage distillation.} 
    We evaluate under various challenging scenes . 
    Our three-stage pipeline consistently delivers more accurate and structured semantic segmentation.
  }
    \label{fig:more}
\end{figure*}
\subsubsection{Knowledge Distillation for Autonomous Driving Perception}

In autonomous driving perception, the high cost and complexity of multi-sensor fusion models pose challenges to scalability and real-time deployment. To address this, knowledge distillation (KD)~\cite{lidar2map2023,chen2023bevdistill,zhao2023simdistill,xu2024sckd,wang2024sparsekd,wang2023distillbev,zhang2023pointdistiller} has emerged as an effective technique for transferring knowledge from large, high-capacity teacher models to lightweight student models.
From the BEV (Bird's-Eye View) perspective, methods like BEVDistill~\cite{chen2023bevdistill} propose BEV-aware distillation to boost performance of monocular BEV perception. In 3D object detection, approaches such as UniDistill~\cite{zhou2023unidistill} achieve cross-modal distillation between LiDAR and camera-based networks, improving 3D detection accuracy without increasing inference complexity.
In the HD map construction domain, MapDistill~\cite{hao2024mapdistill} introduces a structured distillation strategy that transfers low-level geometric features, high-level semantics, and spatial dependencies from a fused LiDAR-camera teacher to an image-only student. This design enables efficient and accurate online map generation using only camera inputs.

\subsection*{Appendix C: Comparing Two-Stage and Our Three-Stage Distillation Frameworks via Visualizations}
\begin{algorithm}[tb]
\caption{Token-Guided 2D Patch Distillation (TGPD)}
\label{alg:tgpd}
\textbf{Input:} $F_{\mathrm{S}}, F_{\mathrm{T}}, F_{\mathrm{C}}$ \\
\textbf{Output:} $\mathcal{L}_{\text{TGPD}}$
\begin{algorithmic}[1]
\STATE $P \gets \text{PatchEmbed}(F)$
\STATE $g \gets \text{AvgPool}(F)$
\STATE $E \gets [g; P]$
\STATE $A^{\mathrm{S}}, A^{\mathrm{T}}, A^{\mathrm{C}} \gets \text{Attn}(E_{\mathrm{S}}, E_{\mathrm{T}}, E_{\mathrm{C}})$
\STATE $\mathcal{L}_{\mathrm{KL}}^{\mathrm{T}} \gets KL(A^{\mathrm{S}} \| A^{\mathrm{T}})$
\STATE $\mathcal{L}_{\mathrm{KL}}^{\mathrm{C}} \gets KL(A^{\mathrm{S}} \| A^{\mathrm{C}})$
\STATE $\mathcal{L}_{\mathrm{MSE}}^{\mathrm{T}} \gets \|E_{\mathrm{S}} - E_{\mathrm{T}}\|_2^2$
\STATE $\mathcal{L}_{\mathrm{MSE}}^{\mathrm{C}} \gets \|E_{\mathrm{S}} - E_{\mathrm{C}}\|_2^2$
\STATE $\mathcal{L}_{\text{TGPD}} =  \mathcal{L}_{\mathrm{KL}}^{\mathrm{T}} +  \mathcal{L}_{\mathrm{KL}}^{\mathrm{C}} + \mathcal{L}_{\mathrm{MSE}}^{\mathrm{T}}+\mathcal{L}_{\mathrm{MSE}}^{\mathrm{C}}$
\STATE \textbf{return} $\mathcal{L}_{\text{TGPD}}$
\end{algorithmic}
\end{algorithm}

To further illustrate the advantages of our proposed three-stage distillation framework, we conduct a comparative analysis with the conventional two-stage distillation strategy.
\appendix

In the traditional two-stage setting, only the teacher and student models are involved. The teacher, equipped with full-modal inputs and stronger representational capacity, provides rich but often complex supervision signals. The student, however, has limited input modalities (e.g., camera-only) and reduced capacity, making it difficult to effectively absorb the teacher’s supervision. As a result, the student often struggles to learn fine-grained structures, leading to suboptimal predictions—particularly for small objects and boundary details.

In contrast, our three-stage framework introduces a coach model as an intermediate bridge. The coach shares similar input modalities with the student and incorporates a pseudo-LiDAR module and HD map pretraining to learn geometry-aware and map-enhanced features. This design ensures that the student receives not only softened but also structurally aligned guidance, improving both convergence and generalization.

\begin{algorithm}[tb]
\caption{Masked Semantic Response Distillation (MSRD)}
\label{alg:msrd}
\textbf{Input:} Logits $S_S, S_T, S_C$ and mask $M$ \\
\textbf{Output:} Distillation loss $\mathcal{L}_{\text{MSRD}}$
\begin{algorithmic}[1]
\STATE Extract valid regions: $\tilde{S} \gets S_S[M], \tilde{T} \gets S_T[M], \tilde{C} \gets S_C[M]$
\STATE Apply sigmoid: $P_T = \sigma(\tilde{T}), P_C = \sigma(\tilde{C})$
\STATE Compute BCE: $\mathcal{L}_T = \text{BCE}(\tilde{S}, P_T), \mathcal{L}_C = \text{BCE}(\tilde{S}, P_C)$
\STATE \textbf{return} $\mathcal{L}_{\text{MSRD}} = \mathcal{L}_T + \mathcal{L}_C$
\end{algorithmic}
\end{algorithm}
We illustrate the advantage of our three-stage framework through representative and challenging scenarios in Figure~\ref{fig:more}, including urban intersections, occluded environments, and complex road geometries. Compared to the two-stage baseline, which often fails to capture key scene boundaries and misses small-scale objects, the three-stage student consistently produces sharper and more semantically complete predictions. This highlights the critical role of the coach in translating and simplifying the teacher’s knowledge into a form that the student can more effectively utilize, demonstrating improved robustness and generalization across diverse conditions.

\subsection*{Appendix D: Algorithm Explanation of the Training Procedure}

Our training process is outlined in Algorithm~\ref{alg:tcs_training}. It includes three main steps: teacher forward, coach forward, and student forward. In each step, we pass images through encoders, convert features into the bird's-eye view (BEV), and then decode them into semantic outputs.

After these forward passes, we compute three types of loss:

\begin{itemize}
    \item \textbf{Base loss:} This is a supervised loss using ground truth maps to guide training.
    \item \textbf{Feature-level loss:} This compares BEV features and attention weights. We use TGPD  to compute the difference as shown in Algorithm~\ref{alg:tgpd}.
    \item \textbf{Logit-level loss:} This compares the final semantic predictions using MSRD in Algorithm~\ref{alg:msrd}. It focuses only on valid foreground areas.
\end{itemize}

By combining these losses, the student learns more effectively. The coach helps by simplifying the teacher’s knowledge. This makes the student more stable and accurate.

\begin{table}[H]
\centering
\setlength{\tabcolsep}{10pt}
\begin{tabular}{lccc}
\hline
\textbf{Parameter} & \textbf{Teacher} & \textbf{Coach} & \textbf{Student} \\
\hline
$\alpha_1$ (dist)      & 0.1   & 0.1   & 0.1 \\
$\alpha_2$ (dir) & 0.1   & 0.1   & 0.1 \\
$\beta_1$ (BEV)       & -   & -  & 0.5 \\
$\beta_2$ (Output)    & -   & -   & 0.5 \\
\hline
\end{tabular}
\caption{\small Loss scaling factors used during the training of the teacher, coach, and student models.}
\label{table:loss}
\end{table}
We train our model on the \textbf{nuScenes} dataset , using a BEV range of $60\,\text{m} \times 30\,\text{m}$. The voxel resolution is set to $0.15\,\text{m}$. Input images are resized to $128 \times 352$. We train for 30 epochs with a batch size of 8, using 4 GPUs and 20 workers.

We use the Adam optimizer with a learning rate of $5 \times 10^{-4}$, weight decay of $1 \times 10^{-7}$, and gradient clipping at 5.0. The loss scaling factors are set in Table \ref{table:loss}, These settings are chosen to promote stable convergence and effective multi-task learning.

\bibliography{aaai2026}

\end{document}